\documentclass[conference]{IEEEtran}
\IEEEoverridecommandlockouts
\usepackage{cite}
\usepackage{amsmath,amssymb,amsfonts}
\usepackage{algorithmic}
\usepackage{graphicx}
\usepackage{textcomp}
\usepackage{xcolor}
\usepackage{algorithmic}
\usepackage[labelsep=period]{caption}
\usepackage{cuted}
\usepackage{array}
\usepackage{capt-of}
\usepackage[caption=false,font=footnotesize]{subfig}
\usepackage{comment}
\usepackage{caption}
\usepackage{lineno}
\usepackage{cleveref}
\usepackage{multirow}
\usepackage{floatrow}
\usepackage{placeins}

\crefname{equation}{Eq.}{Eq.} 
\crefrangelabelformat{equation}{(#3#1#4)--(#5#2#6)}
\crefname{figure}{Fig.}{Fig.}

\def\BibTeX{{\rm B\kern-.05em{\sc i\kern-.025em b}\kern-.08em
    T\kern-.1667em\lower.7ex\hbox{E}\kern-.125emX}}

\newcolumntype{P}[1]{>{\centering\arraybackslash}p{#1}}

\makeatletter
\newcommand{\linebreakand}{%
  \end{@IEEEauthorhalign}
  \hfill\mbox{}\par
  \mbox{}\hfill\begin{@IEEEauthorhalign}
}
\makeatother

\begin{document}

\title{Multi-Task Wavelength-Multiplexed Reservoir Computing Using a Silicon Microring Resonator\\

\thanks{This work has received funding by Vetenskapsrådet (BRAIN, grant n. 2022-04798) and Villum Founden (OPTIC-AI VIL29334, and VI-POPCOM 54486).}
}

\author{\IEEEauthorblockN{Bernard J. Giron Castro}
\IEEEauthorblockA{\textit{DTU Electro} \\
\textit{Technical University of Denmark}\\
2800 Kongens Lyngby, Denmark \\
bjgca@dtu.dk}
\and
\IEEEauthorblockN{Christophe Peucheret}
\IEEEauthorblockA{\textit{CNRS, UMR6082 - FOTON} \\
\textit{Univ Rennes}\\
22305 Lannion, France \\
christophe.peucheret@univ-rennes.fr}
\linebreakand 
\IEEEauthorblockN{Darko Zibar}
\IEEEauthorblockA{\textit{DTU Electro} \\
\textit{Technical University of Denmark}\\
2800 Kongens Lyngby, Denmark \\
dazi@dtu.dk}
\and 
\IEEEauthorblockN{Francesco Da Ros}
\IEEEauthorblockA{\textit{DTU Electro} \\
\textit{Technical University of Denmark}\\
2800 Kongens Lyngby, Denmark \\
fdro@dtu.dk}
}

\maketitle

\begin{abstract}
Among the promising advantages of photonic computing over conventional computing architectures is the potential to increase computing efficiency through massive parallelism by using the many degrees of freedom provided by photonics. Here, we numerically demonstrate the simultaneous use of time and frequency (equivalently wavelength) multiplexing to solve three independent tasks at the same time on the same photonic circuit. In particular, we consider a microring-based time-delay reservoir computing (TDRC) scheme that simultaneously solves three tasks: Time-series prediction, classification, and wireless channel equalization. The scheme relies on time-division multiplexing to avoid the necessity of multiple physical nonlinear nodes, while the tasks are parallelized using wavelength division multiplexing (WDM). The input data modulated on each optical channel is mapped to a higher dimensional space by the nonlinear dynamics of the silicon microring cavity. The carrier wavelength and input power assigned to each optical channel have a high influence on the performance of its respective task. When all tasks operate under the same wavelength/power conditions, our results show that the computing nature of each task is the deciding factor of the level of performance achievable. However, it is possible to achieve good performance for all tasks simultaneously by optimizing the parameters of each optical channel. The variety of applications covered by the tasks shows the versatility of the proposed photonic TDRC scheme. Overall, this work provides insight into the potential of WDM-based schemes for improving the computing capabilities of reservoir computing schemes.
\end{abstract}

\begin{IEEEkeywords}
Reservoir computing, neuromorphic photonics, parallel computing, wavelength division multiplexing.
\end{IEEEkeywords}

\section{Introduction}
The growth of computing demanding applications is pushing the design of novel hardware accelerators with higher power efficiency and a boost in computing capabilities \cite{Dabos:22}. Very high computational resources are essential for large artificial intelligence systems and deep neural networks (NN). Henceforth, the boom of such systems in recent years has only driven further the research on more advanced computational paradigms \cite{Dabos:22, Huang2022}. As Moore's law becomes more challenging to extend in the electronics domain, alternative hardware platforms are under investigation for providing the required computing power \cite{Huang2022}. Photonic computing has emerged as an interesting alternative for delivering the required computational resources in the upcoming years \cite{Huang2022, shastri2021photonics}. Current systems may benefit from the theoretical advantages of light over electronics, such as higher bandwidth and low loss inter-connectivity \cite{shastri2021photonics}. The translation of well-developed multiplexing technologies from optical communications to photonic computing offers the potential to design more advanced parallel computing schemes \cite{Yunpig2023}.

Within machine learning schemes and, more specifically, the family of recurrent NNs, we focus our attention on reservoir computing (RC). In this type of NN, the input data space is mapped to a higher dimensional one through nodes interconnected with random and untrained weights. The nonlinearity given by the complex dynamics of the reservoir layer must ensure the linear separability of the state of the nodes at the output layer. This allows RC to only require linear regression training of its output layer. The other essential feature of RC is the so-called fading memory, which allows the NN to buffer the most significant information of past inputs in a finite interval \cite{Cucchi_2022}. RC schemes have shown good performance in classification, speech recognition, channel equalization, and forecasting applications \cite{Argyris2022}. 

The properties of RC have attracted attention for the research of its physical implementation on different technologies, including photonics \cite{Cucchi_2022, Huang2022}. Some photonic RC schemes require the physical implementation of each node by its individual nonlinear photonic device \cite{Vandoorne2014, Mesaritakis:13}. A more attractive RC approach in terms of requiring a reduced number of physical nonlinear nodes is time-delay RC (TDRC), first investigated in \cite{Appeltant2011-pj}.  The hardware is minimized by multiplexing in time the (virtual) nodes. The response of each node is processed, one at a time, in a single physical nonlinear node \cite{Hulser:22}. Henceforth, much research on photonic RC has been focused on TDRC schemes as they appear to be more practical experimentally. Photonic TDRC has been implemented with photonics using a variety of technologies as the nonlinear node: Through the dynamics of laser devices in \cite{Bueno:17, 8758193, Skontranis_2023}, using Mach-Zehnder modulators \cite{Paquot2012}, and more recently, using silicon microring resonators (MRRs) \cite{Donati:22, GironCastro:24}.

An MRR comprises an optical waveguide forming a closed loop in which resonance occurs when its optical path length is a multiple of the wavelength of the input signal \cite{Bogaerts2012}. It is usually coupled to one (all-pass configuration, two ports) or two (add-drop configuration, four ports) bus waveguides. Further details about silicon MRRs and the effects that are exploited as the main source of nonlinearity in the investigated TDRC scheme are presented in \cref{section2,section4}. A TDRC based on an add-drop MRR with an external waveguide connecting the two bus waveguides to provide delayed feedback of the input data was recently investigated in \cite{Donati:22, GironCastro:24}. It demonstrated to highly enhance the fading memory of the system. In \cite{Donati:22, GironCastro:24}, only one optical channel with a wavelength detuned from a single resonance of the MRR cavity is used. This limits the TDRC to process and solve a single task. Hence, we focus on the potential of using photonic multiplexing techniques to extend the computing capabilities of current MRR-based TDRC systems.  

In optical communications, different data streams can be transmitted over multiple optical channels through the same medium using wavelength-division multiplexing (WDM). In this technique, each stream of data is allocated a different wavelength within the available spectral bandwidth, increasing the transmission capacity. WDM has long been a key technology in the backbone of optical fiber communications \cite{mukherjee2000wdm}. 

Nowadays, WDM has also proven to be feasible for photonic parallel computing as multiple computing tasks can be simultaneously addressed on different optical channels \cite{Yunpig2023, DuanChenLin2023, Yang:12, Gooskens:22, 10.1063/5.0158939}. Reconfigurable photonic circuits, photonic tensor cores, convolutional, feedforward, and recurrent photonic NNs are among the applications of WDM in photonic computing \cite{Yunpig2023}. In \cite{DuanChenLin2023}, different types of classification tasks are addressed in parallel by a multi-wavelength diffractive deep NN modulated on the visible part of the frequency spectrum. A WDM-based on-chip optical signal processor capable of performing matrix-vector multiplication is demonstrated in \cite{Yang:12}.

Previous studies \cite{Gooskens:22, 10.1063/5.0158939} have applied wavelength multiplexing to RC, demonstrating the possibility of addressing the same task through multiple channels. However, these studies relied on using physical RC nodes, i.e. providing lower scalability than the TDRC paradigm, and only addressed solving one specific task.

In this work, we triple the computing power of MRR-based TDRC schemes by using three of the resonances of a single add-drop MRR. Each optical channel in parallel addresses a different and independent task. The three tasks cover a diversity of TDRC applications: Classification, time-series prediction, and wireless channel equalization. They are solved simultaneously with a good performance regarding their metrics. By adding the wavelength dimension for multiplexing to the implicit time multiplexing inherent in TDRC, the proposed WDM-TDRC offers a higher computing potential than other TDRC schemes that solve just a single task at a time.

\section{Fundamentals of Silicon Microring Resonators in an Add-drop Configuration}\label{section2}

\begin{figure}[!h]
\centering
\includegraphics[scale=0.25] {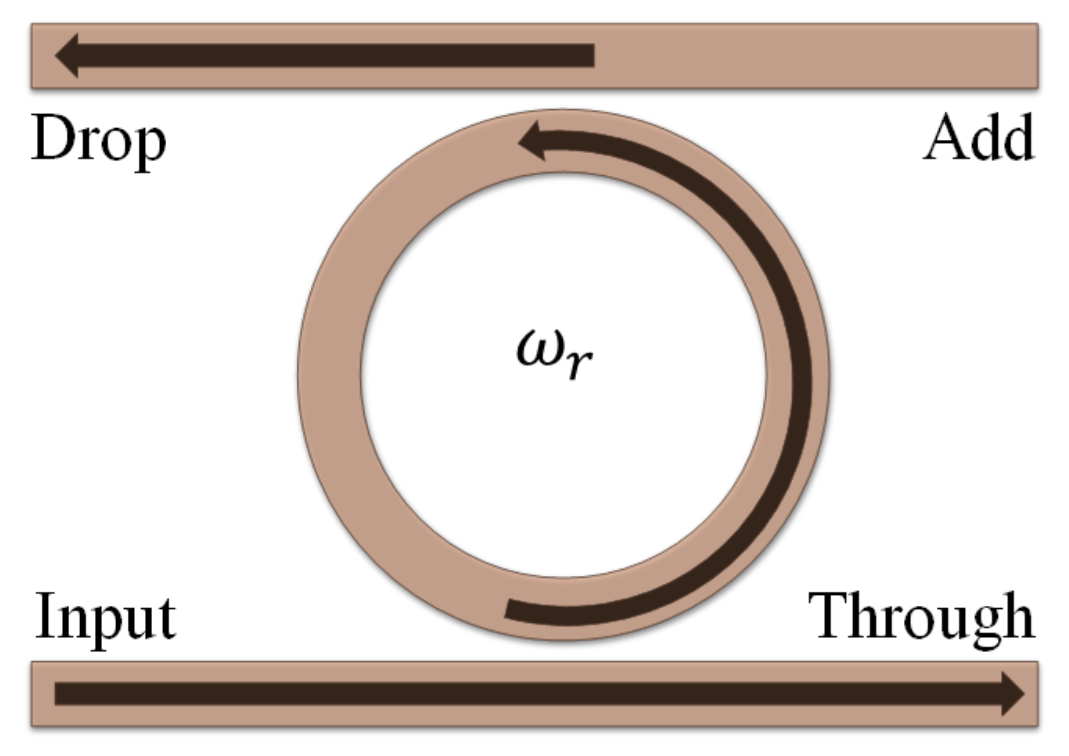}
\caption{Add-drop Microring resonator.}
\label{fig1}
\end{figure}

\begin{figure}[!h]
\centering
\includegraphics[scale=0.33] {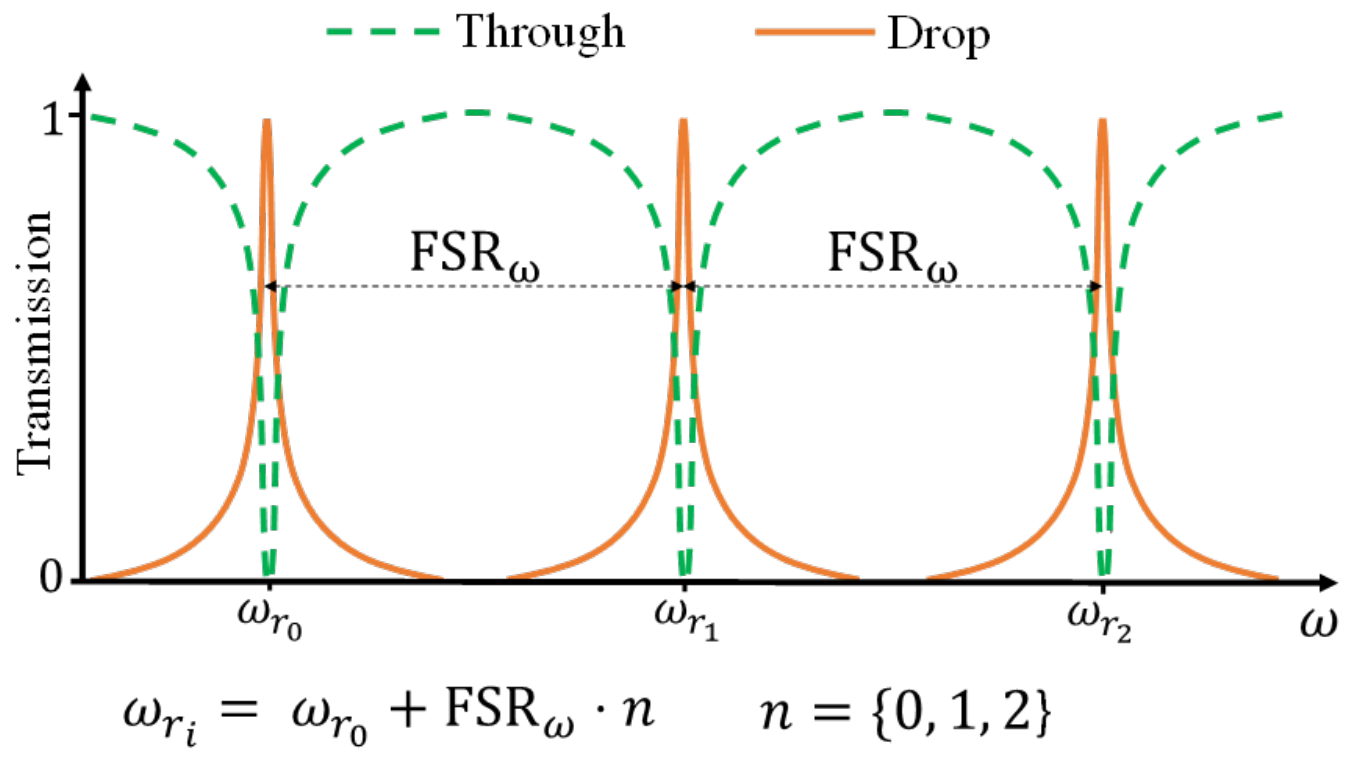}
\caption{Typical linear transmission response of an add-drop MRR: To the through port in green, to the drop port in orange.}
\label{fig2}
\end{figure}

We consider an MRR in add-drop configuration as shown in \cref{fig1} for a silicon on insulator platform (SOI). The high refractive index contrast between the silicon waveguide and the silica cladding considered in our model provides a solid confinement of a single optical mode \cite{Bogaerts2012}. Light propagated from the input port to the through port at a frequency matching a resonance frequency of the cavity, $\omega_r$, is coupled into the second bus waveguide and propagated towards the drop port. An example of the typical transmission response of an MRR is shown in \cref{fig2}. The angular resonance frequency of the $l^{\textrm{th}}$ resonance contained by an MRR is dependent on its radius $R$ and the effective refractive index of the propagated mode, $n_{\textrm{eff}}$ as expressed in \cref{eq1}\cite{Bogaerts2012}. $c$ is the speed of light in vacuum.  

\begin{equation}\label{eq1}
    \omega_r = {\frac{c}{n_{\textrm{eff}}R}}\cdot l, \qquad l = {1, 2, 3...}
\end{equation}
\vspace{-0.001cm}

The wavelength range between two resonances is commonly known as the free spectral range (FSR). In terms of a resonance wavelength and the group index $n_g$, it is determined as \cite{Bogaerts2012}:

\begin{equation}\label{eq2}
    \textrm{FSR}_\lambda \approx \frac{{\lambda}^2}{2\pi R n_g}.
\end{equation}

Nonlinear dynamics can be triggered in a silicon MRR \cite{Bogaerts2012}. We introduce in it an optical signal with a frequency $\omega_i$, close to $\omega_r$. The optical mode propagating through the MRR with a modal amplitude 
 $a$, generates an excess of free carriers in the cavity. Two-photon absorption (TPA), along with Free-carrier dispersion (FCD) and thermo-optic (TO) effects rise, increasing the loss rate in the cavity due to power absorption and dissipated heat. However, as $a$ is reduced, FCD and the TO effect decrease. In turn, their caused losses are diminished. Hence, $a$ increases, and the cycle repeats \cite{Bogaerts2012, Borghi2016, PhysRevA.87.053805}. The oscillation of $a$ within the cavity generates a self-pulsing effect \cite{PhysRevA.87.053805}. The resulting transition between linear and nonlinear regimes is essential for MRR-based TDRC systems \cite{Donati:22, GironCastro:24}. The modeling of the MRR dynamics is presented in \cref{section4}.

\section{Methodology}\label{section3}
\subsection{Proposed Scheme}

\begin{figure*}[!h]
\centering
    \includegraphics[scale=0.255, trim={0.5cm 0.0cm 0cm 0.25cm}]{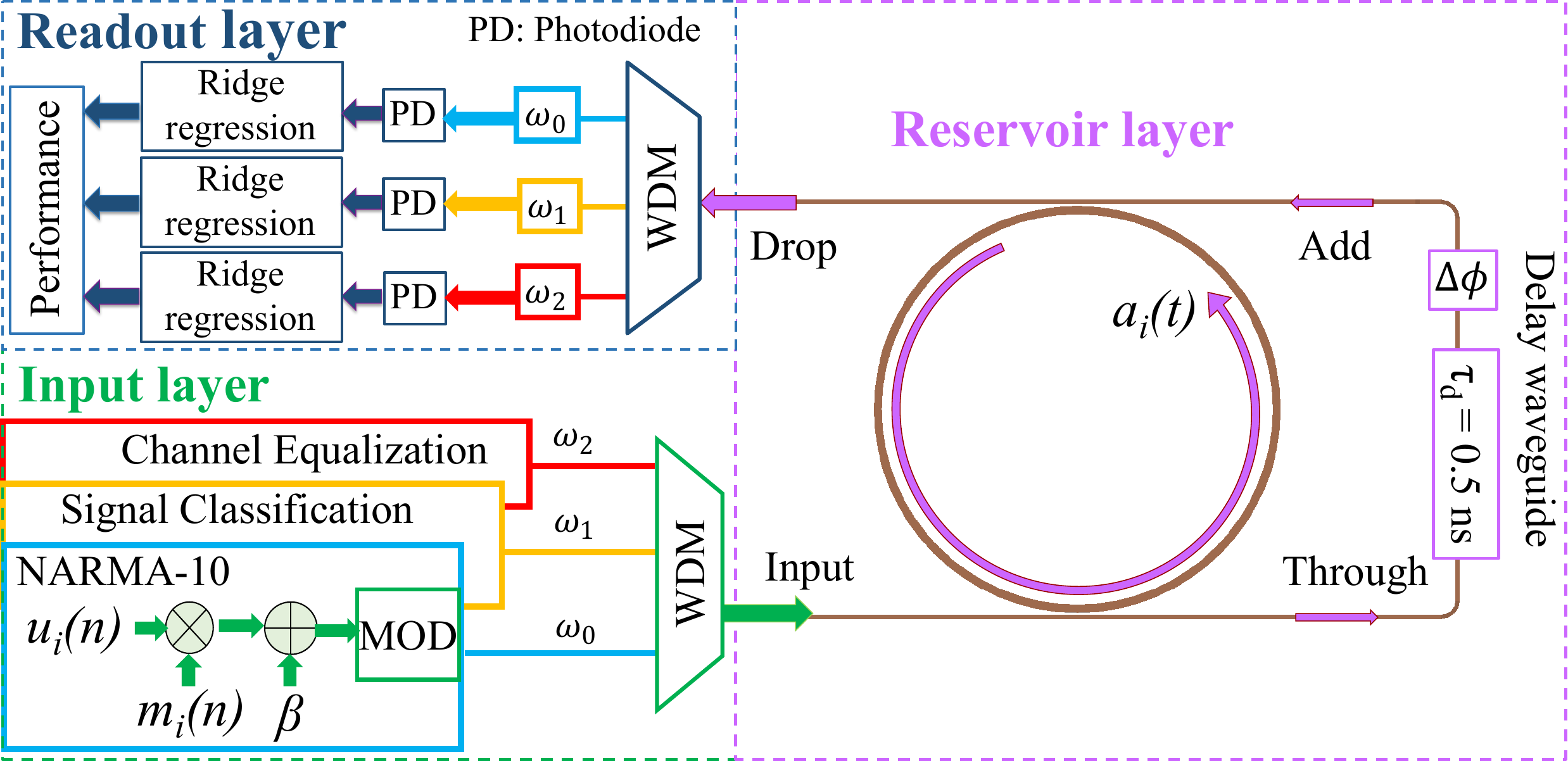}
\captionof{figure}{Proposed TDRC setup scheme solving three independent tasks in parallel through wavelength multiplexing.}

\label{fig3}
\end{figure*}

The setup of the investigated TDRC scheme is shown in Fig. \nolinebreak \ref{fig3}. The three conventional TDRC benchmarking tasks solved simultaneously are described in \cref{subsecIIIB}, including the definition of the input signals and masking procedures. A 1-GBd input symbol sequence is used per task. The input sequence $u_i(n)$ of each task $i$, is multiplied by its respective masking signal $m_i(n)$ for $N = 50$ virtual nodes and an optimized task-independent bias $\beta$ is added to their product.

\begin{figure}[!t]
\includegraphics[scale=0.33] {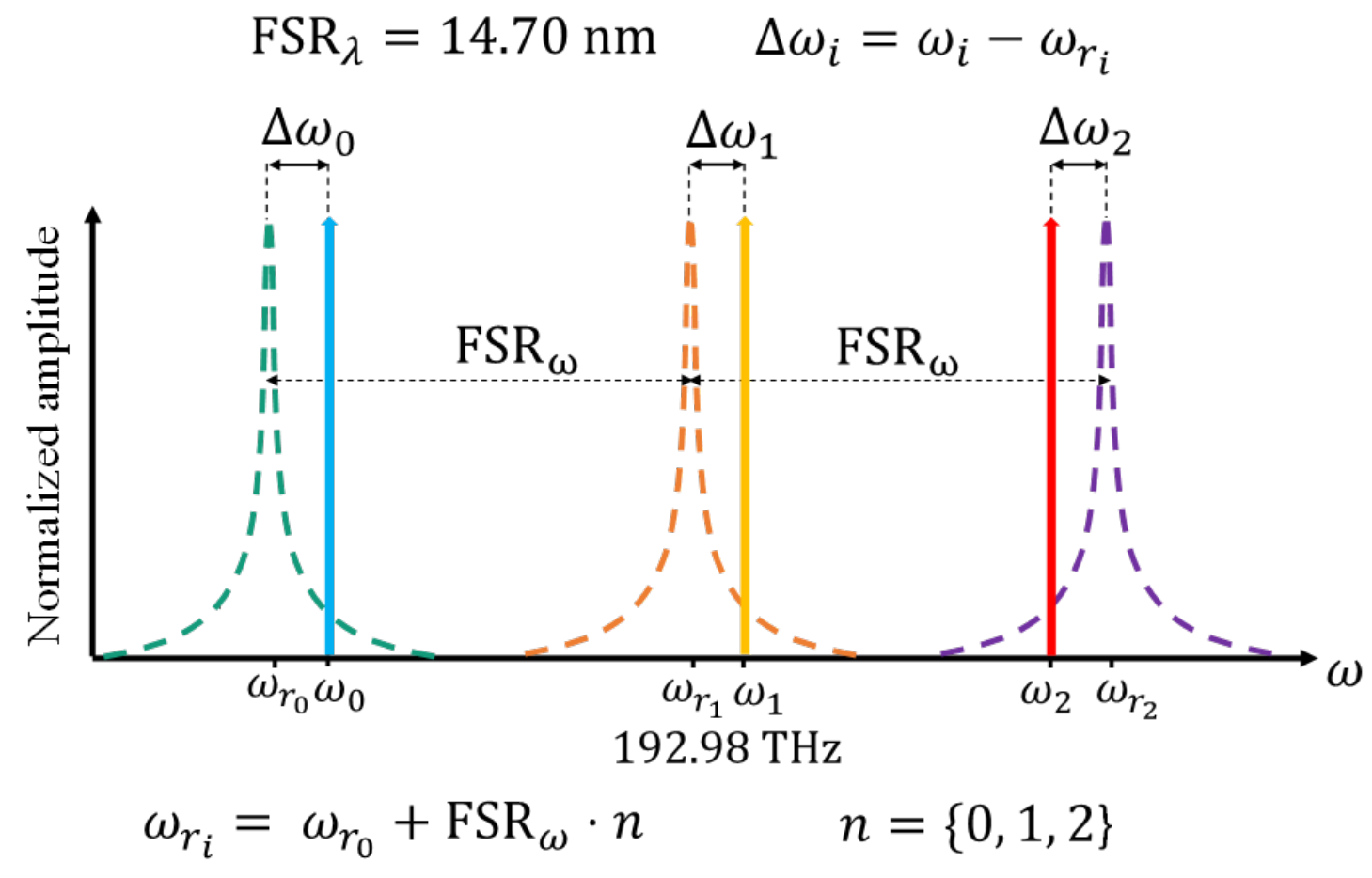}
\caption{Frequency allocation used in this work.}
\label{fig4}

\end{figure}

Each of the resulting signals modulates the intensity of its respective optical channel before being wavelength-multiplexed and injected into the MRR (modeled as in Section \ref{section4}). The MOD block in \cref{fig3} consists of a square root factor to account for the optoelectronic conversion of the signal followed by an assumed linear modulation of each optical channel. The corresponding frequency allocation of the respective optical channel of each task is shown in Fig. \ref{fig4}. We connect the input and add ports of the MRR to enhance the RC memory as in \cite{GironCastro:24, Donati:22}. The electric fields of the $i^{th}$ carrier at the add ($E_{\textrm {add}_i}$) and drop ($E_{\textrm {drop}_i}$)  ports of the MRR can be expressed as:

\begin{equation}
    E_{\textrm {add}_i}(t) = \kappa e^{-i\phi}\left[E_{\textrm {in}_i}(t-\tau_\textrm d )+\frac{1}{\tau_\textrm c}a_i(t-\tau_\textrm d )\right],
\end{equation}

\begin{equation}
    E_{\textrm {drop}_i}(t) = \frac{1}{\tau_\textrm c}a_i(t)E_{\textrm {in}_i}(t)+E_{\textrm {add}_i}(t), 
\end{equation}

\noindent where $E_{\textrm {in}_i}$ is the $i^{th}$ electric field at the input port, $a_i(t)$ is the modal amplitude of the $i^{th}$ optical channel and $\tau_\textrm d = \nolinebreak 0.5$ \nolinebreak ns is the time delay added by the external feedback. $1/\tau_\textrm c$ is the loss rate of the cavity due to the coupling with the bus waveguides. $\kappa = 0.95$ is the coupling factor of the external waveguide and $\phi$ is the total phase shift of the optical signals when propagated through the external waveguide. It is defined as:

\begin{equation}
    \phi= \frac{2\pi \tau_\textrm d c}{\lambda_\textrm p} + \Delta\phi.
\end{equation}

There, $\Delta\phi = 0.15 $ is an external phase shift adjusted to achieve good performance on every task studied in this work. At the drop port, the three WDM signals are demultiplexed and detected by individual photodiodes. For the training of each RC task, we use ridge regression with a regularization parameter $\Lambda$ = $0.5\times10^{-10}$, which was adjusted in terms of best performance of the tasks. To simplify the simulation process, the system uses the same data lengths for the three tasks, 10000 symbols for training and 10 different subsets of 10000 symbols for testing. The mathematical model of the system used for its simulation is further described in section \ref{section4}.

\subsection{RC Tasks}\label{subsecIIIB}

The potential for WDM to multiply the computing capabilities of the TDRC scheme has been tested by implementing three independent and substantially different tasks, each modulated on one of the three optical channels: NARMA10 time-series prediction, signal classification (SC),  and nonlinear channel equalization. Their descriptions are taken from \cite{Paquot2012}.

\subsubsection{NARMA-10 Time-series Prediction}

The discrete-time tenth-order nonlinear auto-regressive moving average (NARMA-10) is a one-step ahead time-series prediction in which the target of the RC is the function expressed as:

\begin{equation}
\label{eq6}
\begin{aligned}
    y(n+1) = 0.3y(n) + 0.05y(n)\left[\sum_{i=0}^9 y(n-i)\right] \\ + 1.5u(n-9)u(n) + 0.1 . 
\end{aligned}
\end{equation}

The sequence $u_0(n)$ is generated from a uniform distribution over the interval [0.0, 0.5]. The mask $m_0(n)$ is taken from a uniform distribution in the range [0.0, +1.0]. The performance is measured by calculating the normalized mean square error (NMSE) between the predicted and target sequences as defined in \cref{eq7}. $\hat{y}(n)$ is the predicted sequence and $y(n)$ is the target sequence. $\sigma_{y}^2$ is the standard deviation of $y(n)$ and $L_{\textrm {data}}$ its length. 

\begin{equation}\label{eq7}
    \textrm {NMSE} = \frac{1}{L_{\textrm {data}}}\frac{\sum_{n} \left( \hat{y}(n) - y(n) \right)^2}{\sigma_{y}^2}.   
\end{equation}

\subsubsection{Signal Classification}

In this task, the target of the system is to distinguish between a square and a sine waveform. Therefore, the input sequence $u_1(n)$ is a stream of random sequences of sine and square waves that are discretized over 12 points per period. $m_1(n)$ is taken from a uniform distribution over the interval [0, +1]. When the signal is classified as a square wave, the value of the prediction $\hat{y}(n)$ is 1, and 0 when the signal is a sine. The metric of performance for this task is the classification accuracy of the system. It is obtained by dividing the number of accurate predictions by their total number. 

\subsubsection{Wireless Channel Equalization}

This task evaluates the performance of the RC when reconstructing the original input signal after it has been propagated through a wireless channel model affected by noise and nonlinear distortion. The channel is modeled as a linear system which is followed by a combination of second-order and third-order nonlinear distortion. The original signal, $d(n)$ is an independent, identically distributed random sequence with values $\{3, -1, +1, +3\}$. The output of the linear system $q(n)$ is mathematically described by the following expression:

\begin{equation}
\begin{split}
q(n) = 0.08d(n+2) - 0.12d(n+1) + d(n)  \\ + 0.18d(n-1)  - 0.1d(n-2) + 0.091d(n-3)  \\ - 0.05d(n-4) + 0.04d(n-5) + 0.03d(n-6)  \\ + 0.01d(n-7).
\end{split}
\end{equation}\label{eq8}

The input sequence of the reservoir $u(n)$ is the output of the system after $q(n)$ is affected by the nonlinear distortions and pseudo-random additive Gaussian noise with zero mean, $v(n)$. This process can be expressed as:

\begin{equation}
    u(n) = q(n) + 0.036q(n)^2 - 0.011q(n)^3 + v(n).
\end{equation}\label{eq9}

At the input layer of the TDRC, we introduce a biased version of the input: $u(n) + 30$. The masking sequence for this task, $m_2(n)$ is uniformly distributed over the interval $[-1, +1]$. The output of the RC is rounded to the closest symbol in $\{3, -1, +1, +3\}$ to form the reconstructed signal $\hat{y}(n)$. Then, the target sequence, i.e, the original signal, is shifted in time by 2, so that the TDRC is trained to minimize the square error as: $(\hat{y}(n) - d[n-2])^2$. For this task, we calculate the symbol error ratio (SER) between the original and reconstructed signals at a signal-to-noise ratio (SNR) of 32 dB. 

\section{Modeling of the Proposed WDM-TDRC}\label{section4}
\subsection{Physical Model}
Temporal coupled-mode theory (TCMT) is used to model an add-drop SOI MRR with multiple input WDM signals in its cavity \cite{GironCastro:24,Borghi2016}. We account for the rate change of $a_i(t)$ inside the MRR when we introduce an input optical signal with frequency $\omega_i$ close to its assigned resonance frequency $\omega_{r_i}$. The model includes individual contributions of each optical channel to the total rate of change of the mode-averaged temperature with respect to the environment ($\Delta T$) and the excess free-carrier density generated via TPA ($\Delta N$). The model is defined for an $M$ number of optical channels as:

\begin{equation}\label{eq10}
\frac{\textrm da_i(t)}{\textrm dt} = [j\delta_i(t)-\gamma_i(t)]a_i(t) + j
	\kappa_\textrm c\left[E_{\textrm {in}_i}(t)+E_{\textrm {add}_i}(t)\right],
\end{equation}

\begin{equation}\label{eq11}
\frac{\textrm d\Delta N(t)}{\textrm dt} = -\frac{\Delta N(t)}{\tau_{\textrm {FC}}} + 
        \sum _{i=1}^M \frac{\Gamma_{\textrm {FCA}}c^2 \beta_{\textrm {TPA}}}{2\hbar\omega_pV^2_{\textrm {FCA}}n^2_{\textrm {Si}}} |a_i(t)|^4,
\end{equation}

\begin{equation}\label{eq12}
\frac{\textrm d\Delta T(t)}{\textrm dt} = -\frac{\Delta T(t)}{\tau_{\textrm {th}}} + 
        \frac{\Gamma_{\textrm {th}}}{mc_\textrm p} \left[ \sum _{i=1}^M P_{\textrm {abs}_i}(t)|a_i(t)|^2\right],
\end{equation}

\noindent where $\delta_i(t)$ is the total angular frequency detuning per optical channel including carrier-resonance ($\omega_i - \omega_{r_i}$) detuning as well as TO and FCD-induced detuning as in \cref{eq13}. $\gamma_i(t)$ and $P_{\textrm {abs}_i}(t)$ denote the total losses and power absorbed in the cavity and are defined in \cref{eq14,eq15}, respectively. $\gamma_{\textrm{TPA/FCA}}$ are the losses due to FCA and TPA. $\hbar$ denotes the reduced Planck’s constant. $\tau_{\textrm {FC}}$ is the lifetime of the carriers, $\tau_{\textrm {th}}$ is the heat diffusion time constant, and $m$ is the mass of the MRR. $\Gamma_{\textrm{FCA/th}}$ refer to the FCA and thermal confinement factors. $n_{\textrm {Si}}$, $\beta_{\textrm {TPA}}$ and $c_\textrm p$, are silicon’s refractive index, TPA coefficient, and specific heat, respectively. $V_{\textrm{FCA/TPA}}$ are the FCA and TPA effective volumes, respectively. $\sigma_{\textrm {FCA}}$ is the total FCA cross-section. In \cref{eq13,eq14,eq15}, d$n$/d$N$ and d$n$/d$T$ are the silicon FCD and TO coefficients, respectively. $\alpha$ is the waveguide attenuation and $\kappa_\textrm c = \sqrt{2/\tau_\textrm c}$ is the factor of modal amplitude decay rate due to the coupling between the MRR and the bus waveguides. The values of the optical parameters used in this work are listed in Table \ref{tab1}.

\begin{equation}\label{eq13}
    \delta_i(t) = \omega_i - \omega_{r_i} + \frac{\omega_{r_i}}{n_{\textrm {Si}}}\left(\Delta N(t) 
     \frac{\textrm dn_{\textrm {Si}}}{\textrm dN} + \Delta T(t) \frac{\textrm dn_{\textrm {Si}}}{\textrm dT} \right),
\end{equation}

\begin{equation}\label{eq14}
\begin{aligned}
    \gamma_i(t) &= \frac{c\alpha}{n_{\textrm {Si}}} + \frac{2}{\tau_\textrm c} + \gamma_{\textrm {TPA}_i} + \gamma_{\textrm {FCA}} \\ &= \frac{c\alpha}{n_{\textrm {Si}}} + \frac{2}{\tau_\textrm c} + \frac{\beta_{\textrm {TPA}}c^{2}}{n_{\textrm {Si}}^2V_{\textrm {TPA}}}|a_i(t)|^2 \\ & +\frac{\Gamma_{\textrm {FCA}}\sigma_{\textrm {FCA}}c}{2n_{Si}} \cdot \Delta N(t),
    \end{aligned}
\end{equation}

\begin{equation}\label{eq15}
    P_{\textrm {abs}_i}(t) = \left(\frac{c\alpha}{n_{\textrm {Si}}} + \gamma_{\textrm {TPA}_i} + \gamma_{\textrm {FCA}}\right)|a_i(t)|^2.
\end{equation}

\begin{table}[h!]
    \centering
    \begin{tabular}[c]{|c|c|} 
         \hline
         Parameter & Value\\
         \hline
         $m$   & $1.2\times10^{-11}$ kg \cite{GironCastro:24}    \\$\beta_{\textrm {TPA}}$&   $8.4\times10^{-11}$ m $\cdot$ W$^{-1}$ \cite{VanVaerenbergh:12} \\
         $\tau_{\textrm c}$& $54.7$ ps \cite{GironCastro:24} \\ $\Gamma_{\textrm {FCA}}$  & 0.9996 \cite{VanVaerenbergh:12} \\
         $n_{\textrm {Si}}$ & 3.485 \cite{Johnson:06} \\ $\Gamma_{\textrm {th}}$  & 0.9355 \cite{VanVaerenbergh:12}\\
         $\lambda_{1} $ & $1553.49$ nm  \cite{GironCastro:24}   \\ d$n_{\textrm {Si}}/$d$T$ &  $1.86$ $\times$ 10$^{-4}$ K$^{-1}$ \cite{Johnson:06}\\
         $L$&   $2\pi\cdot7.5$ $\mu$m \cite{GironCastro:24}   \\ d$n_{\textrm {Si}}/$d$N$ & $-1.73$ $\times$ 10$^{-27}$ m$^{-3}$ \cite{VanVaerenbergh:12}\\
         $c_{\textrm p}$& 0.7 J $\cdot$ (g $\cdot$ K)$^{-1} $ \cite{Johnson:06} \\ $\sigma_{\textrm {FCA}}$& 1.0 $\times$ 10$^{-21}$ m$^2$ \cite{VanVaerenbergh:12}\\
         $V_{\textrm {FCA}}$& 2.36 $\mu$m$^{3}$ \cite{VanVaerenbergh:12}\\ $V_{\textrm {TPA}}$& 2.59 $\mu$m$^{3}$ \cite{VanVaerenbergh:12}\\
         $\tau_{\textrm {th}}$ & 50 ns \cite{GironCastro:24}\\
         $\tau_{\textrm {FC}}$ & 10 ns \cite{GironCastro:24}\\
         $\alpha$ & 0.8 dB/cm \cite{GironCastro:24}\\
         \hline
     
    \end{tabular}\par
    \caption{Optical parameters used in the photonic RC simulations.}
    \label{tab1}
\end{table}

\subsection{Simulation Procedure}
The system of differential equations \cref{eq10,eq11,eq12} is normalized and solved using a $4^{\textrm{th}}$ order Runge-Kutta solver with a step $\eta = 2.0$ ps. The symbols of each task input sequence are masked by $m_i(n)$ to obtain a virtual node with a duration of $\theta = \frac{1.0 \textrm {ns}}{N} = 20$ ps. An optimized bias is added to the masked signal and then, the resulting signals of each virtual node are discretized over the solver steps so that there are $K = \frac{\theta}{\eta} = 10$ solver steps per virtual node. The same procedure is applied to the time delay added by the external feedback to determine the number of solver steps required in the TCMT model to account for the delay. The discretization procedure of each electric field to determine the Runge-Kutta solution follows the same procedure described in \cite{GironCastro:24}.

As mentioned before, in the readout layer, we demultiplex in frequency each optical channel. The corresponding value of each $j^{\textrm{th}}$ virtual node for its $\theta$ duration is calculated as the average of the $K$ step values of the Runge-Kutta solution:

\begin{equation}
    E_{\textrm {drop}_i}(n) = \frac{1}{M}\sum_{k=(j-1)M+1}^{jM} \hat{E}_{\textrm {drop}_i}(k).
\end{equation}

The quadratic response of the photodiode is accounted for by calculating the square of $E_{\textrm {drop}}(n)$ for each optical channel:

\begin{equation}
    X_{\textrm {drop}_i}(n) = \lvert E_{\textrm {drop}_i}(n)\rvert^2.
\end{equation}

\section{Results and Discussion}\label{section5}

\begin{figure}[!h]
\includegraphics[scale=0.22, trim={0cm 0cm 0cm 0cm}]{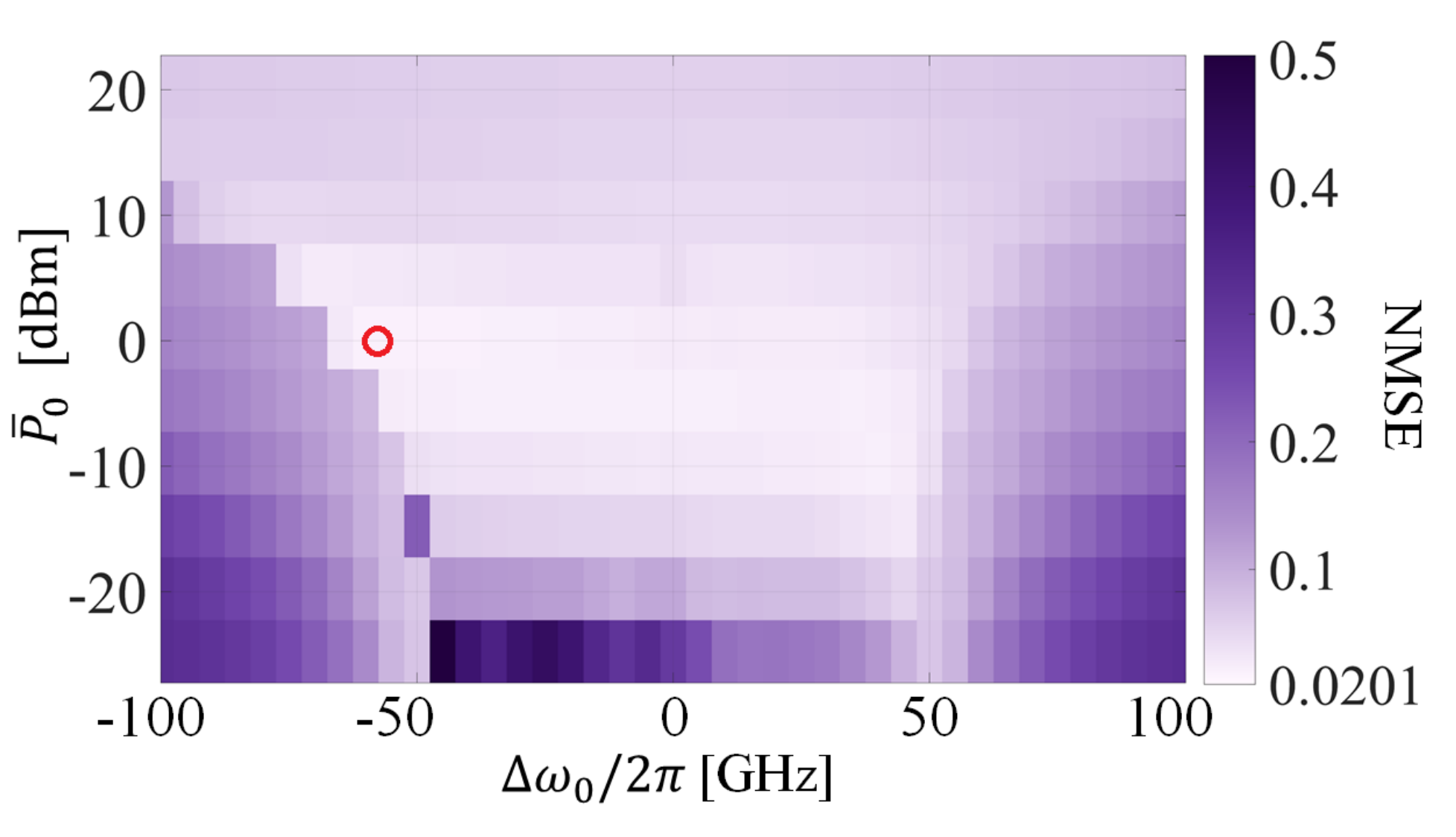}
\caption{NMSE of the NARMA-10 task as a function of $\overline{P}_0$ and $\Delta\omega_0/2\pi$. A red circle marks the best performance.}
\label{fig5}
\end{figure}

\begin{figure}[!h]
\includegraphics[scale=0.22, trim={0cm 0cm 0cm 0.6cm}]{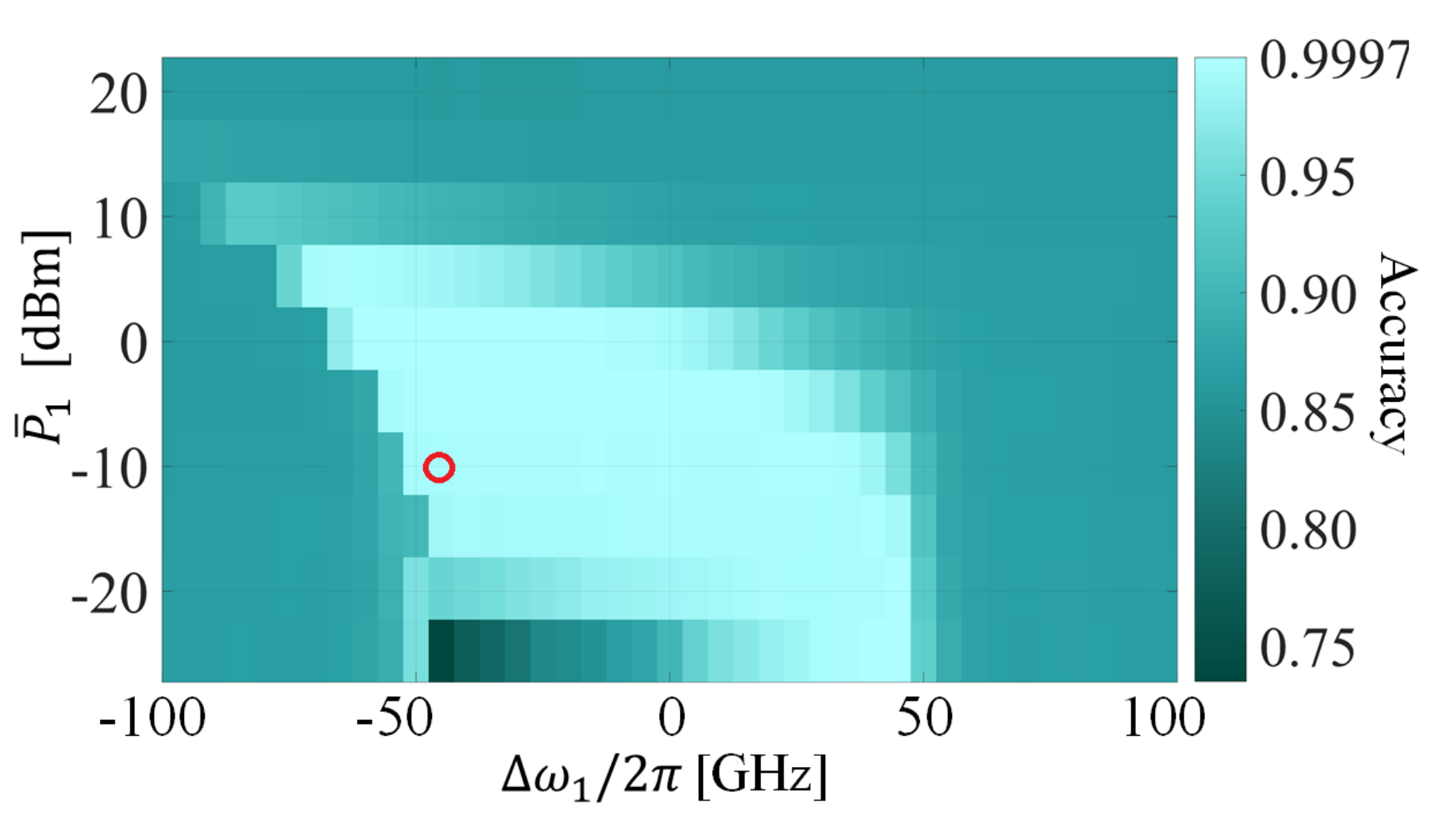}
\caption{Accuracy of the signal classification task as a function of $\overline{P}_1$ and $\Delta\omega_1/2\pi$. A red circle marks the best performance.}
\label{fig6}
\end{figure}

\begin{figure}[!h]
\includegraphics[scale=0.22, trim={0cm 0cm 0cm 0.6cm}]{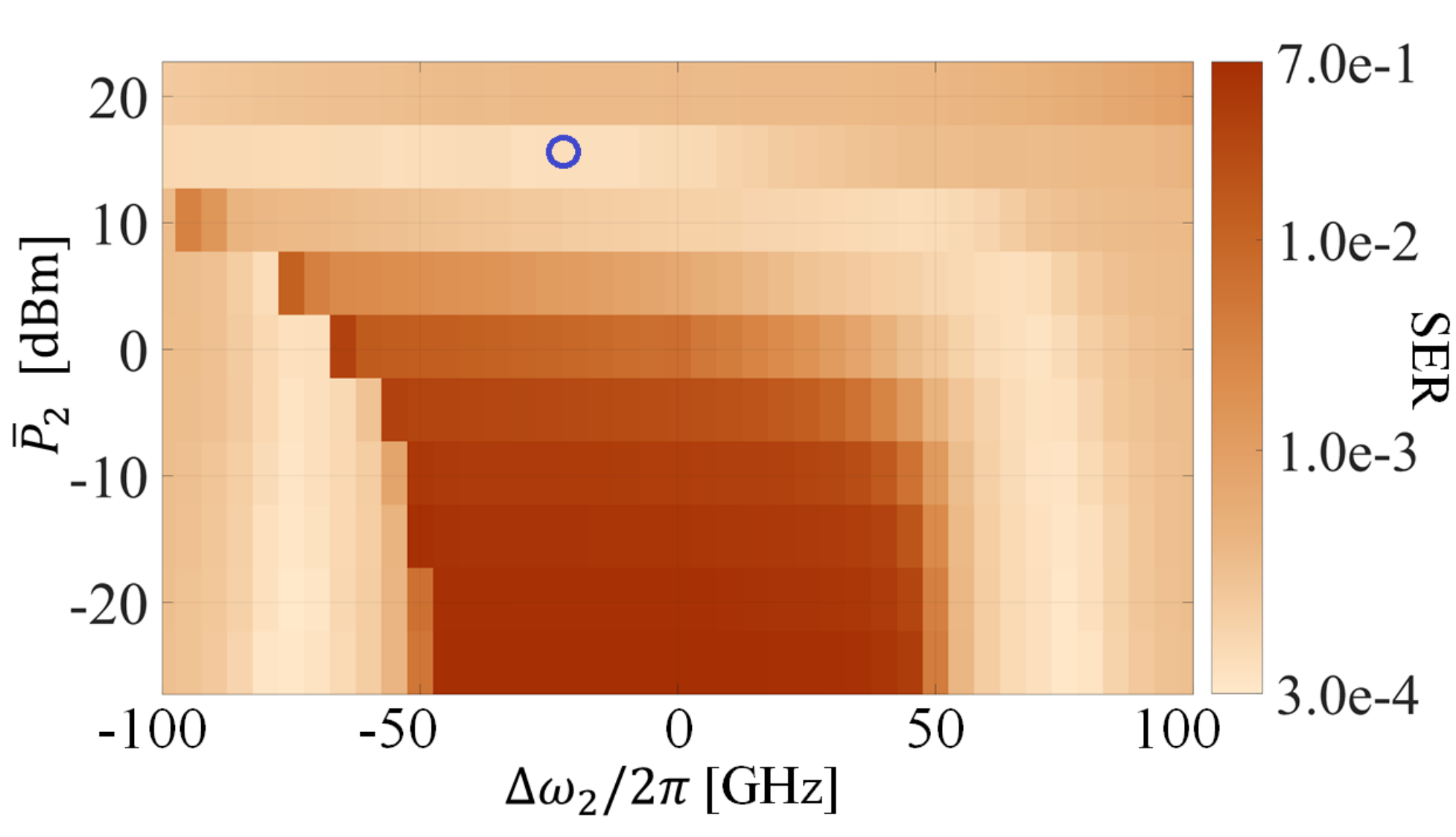}
\caption{SER of the channel equalization task as a function of $\overline{P}_2$ and $\Delta\omega_2/2\pi$. A blue circle marks the best performance.}
\label{fig7}
\end{figure}

 First, we perform the simulation of the system by considering a range of –20 to +25 dBm of total average input power which is split equally between the three optical channels ($\overline{P}_0 = \overline{P}_1 = \overline{P}_2$). For this simulation, we also consider an equal carrier-resonance detuning ($\Delta\omega_0 = \Delta\omega_1 = \Delta\omega_2$) spanning the $\pm$100 GHz range. The results for each task are averaged over 10 different seeds (for the generation of the input signals of each task), and shown in \cref{fig5,fig6,fig7} as a function of $\overline{P}_i$ and $\Delta\omega_i/2\pi$. Minima and maxima values per task metric are also displayed as extreme values of the colorbar legends. The best performance per task is encircled.

FCD and the TO effect depend on the modal amplitude contribution of each optical channel. Therefore, the nonlinear dynamics affecting each optical channel, and in turn, each task, become co-influenced by the other optical channels. Consequently, this conditions the functionality of our system to the existence of common ground of parameters per optical carrier for which all tasks can achieve acceptable performance. In \cref{fig5,fig6,fig7}, we observe that the time-series (NARMA-10) prediction (\cref{fig5}) and classification (\cref{fig6}) tasks share a common area of good performance (low error prediction or high accuracy) which extends over a similar parameter space ($\overline{P}_i$, $\Delta\omega_i$). Regarding the wireless channel equalization task (\cref{fig7}), the parameter space of the lowest SER is located at higher levels of $\overline{P}_i$ than for the NARMA-10 or signal classification tasks and is limited to a smaller input power range. In fact, the parameter space in which the NARMA-10 and signal classification tasks achieve their best performance overlaps the one in which the channel equalization task presents a high SER. This highlights the different computing nature, in terms of memory and nonlinearity, from each task depending on their mathematical definitions. However, this also implies that under the previous simulation conditions, we cannot expect optimum performance simultaneously for the three tasks. 

Nonetheless, even without targeting a complex full 6-D optimization, we still can exploit the results of \cref{fig5,fig6,fig7} as a basis to test that the three tasks can be simultaneously solved with good performance.
Therefore we simulate the system with the carrier power and $\Delta\omega_i/2\pi$ values corresponding to the best observed performance for each task: [$\overline{P}_0 = 0$ dBm, $\overline{P}_1 = -10.0$ dBm, $\overline{P}_2 = 15.0$ dBm], and $\Delta\omega_i$: [$\Delta\omega_0/2\pi = -60$ GHz, $\Delta\omega_1/2\pi = -45$ GHz, $\Delta\omega_2/2\pi = -20$ GHz].
We obtain the performance of each of the tasks when simulating our system under the defined parameters. Then, we compare the performance per task with the results of other photonic TDRC schemes when using the same number of virtual nodes. This comparison is shown in \cref{tab2,tab3,tab4}. 

Under the aforementioned conditions, each solved task suffers a slight performance penalty with respect to their best results (\cref{fig5,fig6,fig7}). Nonetheless, the results are still comparable to previous works while achieving parallel computing of the tasks.
As this system has not been optimized thoroughly, we expect even further improvements in performance per task to be achievable, e.g., with a systematic 6-D optimization. In this work, we have used a limited number of resonances ($M=3$). 

\begin{table}[ht!]
    \centering
    \begin{tabular}[c]{|c|c|P{1.5cm}|P{2.5cm}|} 
         \hline
         NMSE & $N$ & Processing speed & Reference\\
         \hline
         $0.168 \pm 0.015$  & 50 & 0.2 GBd & Paquot et al. \cite{Paquot2012} (Exp., ST)\\
         \hline
         $0.107 \pm 0.012$  & 50 & 0.9 MBd & Vinckier et al. \cite{Vinckier:15} (Exp., ST)\\
         \hline
         $0.062 \pm 0.008$  & 50 & 0.9 MBd & Vinckier et al. \cite{Vinckier:15} (Num., ST)\\
         \hline
         $0.204 \pm 0.026$  & 25 & 1.0 GBd & Donati et al. \cite{Donati:22} (Num., ST)\\
         \hline
         $0.010\pm 0.009$  & 100 & 1.0 GBd & Donati et al. \cite{Donati:22} (Num., ST)\\         
         \hline
         $0.103\pm 0.018$  & 50 & 0.4 GBd & Chen et al. \cite{Chen:19} (Num., ST)\\         
         \hline         
         $0.0151\pm 0.0021$  & 50 & 1.0 GBd & Giron Castro et al.\cite{GironCastro:24} (Num., ST)\\
         \hline         
         $0.0373\pm 0.0021$  & 50 & 1.0 GBd & This work (Num., MT)\\ 
         \hline         
    \end{tabular}\par
    \caption{Performance comparison for the NARMA-10 task. ST: Single task. MT: Multiple tasks (M=3). Exp: Experimental work. Num: Numerical work.}
    \label{tab2}
\end{table}

\begin{table}[H]
    \centering
    \begin{tabular}[c]{|P{2.5cm}|c|P{1.5cm}|P{2.5cm}|} 
         \hline
         Accuracy & $N$ & Processing speed & Reference\\
         \hline
         $\approx100\%$ (NMSE $\approx 1.5\times10^{-3})$ & 50 & 0.2 GBd & Paquot et al. \cite{Paquot2012} (Exp., ST)\\
         \hline
         99.75\%  & 36 & 1.0 GBd & Vandoorne et al. \cite{Vandoorne:08} (Num., ST)\\
         \hline      
         99.1\%  & 50 & 1.0 GBd & This work (Num., MT)\\ 
         \hline         
    \end{tabular}\par
    \caption{Performance comparison for the signal classification task. }
    \label{tab3}
\end{table}

\begin{table}[H]
    \centering
    \begin{tabular}[c]{|c|c|P{1.5cm}|P{2.5cm}|} 
         \hline
         SER & $N$ & Processing speed & Reference\\
         \hline
         $1.3\times10^{-4}$  & 50 & 0.2 GBd & Paquot et al. \cite{Paquot2012} (Exp., ST)\\
         \hline
         $2.0\times10^{-5}$ & 50 & 0.9 MBd & Vinckier et al. \cite{Vinckier:15} (Exp., ST)\\
         \hline
         $4.0\times10^{-6}$  & 50 & 1.0 GBd & Jin et al. \cite{Jin:22} (Num., ST)\\         
         \hline
         $7.0\times10^{-4}$  & 50 & 0.4 GBd & Chen et al. \cite{Chen:19} (Num., ST)\\         
         \hline         
         $4.0\times10^{-5}$  & 50 & 0.8 GBd & Yue et al.\cite{Yue:19} (Num., ST)\\
         \hline         
         $7.0\times10^{-4}$ & 50 & 1.0 GBd & This work (Num., MT)\\ 
         \hline         
    \end{tabular}\par
    \caption{Performance comparison for the wireless channel equalization task at a SNR = 32dB.}
    \label{tab4}
\end{table}
\FloatBarrier
\newpage

However, more resonances could potentially be modulated with additional tasks in future studies, which would further increase the versatility of the system.

\section{Conclusion}

The potential of a WDM silicon MRR-based TDRC scheme to solve simultaneously and with good performance, three conventional TDRC tasks is numerically demonstrated. Here, we limit the analysis to three simultaneous tasks, however higher degree of parallelization could be possible using more MRR resonances. The results indicate the possibility of fine-tuning the power and frequency detuning of each channel, carrying a distinct task, to simultaneously achieve near-optimal performance. The performance of each task is comparable to state-of-the-art results of previous single-task TDRC implementations. Implementing the system on a SOI platform offers the advantage of a more practical integration with the well-established infrastructure of electronics.
\bibliographystyle{ieeetr}
\bibliography{IEEEWCCI2024}
\end{document}